\documentclass{article}

    \usepackage[final]{neurips}

\usepackage[utf8]{inputenc} % allow utf-8 input
\usepackage[T1]{fontenc}    % use 8-bit T1 fonts
\usepackage{hyperref}       % hyperlinks
\usepackage{url}            % simple URL typesetting
\usepackage{booktabs}       % professional-quality tables
\usepackage{amsfonts}       % blackboard math symbols
\usepackage{nicefrac}       % compact symbols for 1/2, etc.
\usepackage{microtype}      % microtypography
\usepackage{xcolor}         % colors
\usepackage[pdftex]{graphicx}
\usepackage{listings}
\usepackage{longtable}

\title{Semantic Code Classification for Automated Machine Learning}

\author{%
  Polina Guseva$^*$ \\
 HSE University\\
  \texttt{paguseva@edu.hse.ru} \\
   \And
   Anastasia Drozdova\thanks{Equal contribution.} \\
   HSE University \\
   \texttt{avdrozdova\_2@edu.hse.ru} \\
   \AND
   Natalia Denisenko \\
   HSE University \\
   \And
   Daria Sapozhnikova \\
   HSE University \\
   \And
   Ivan Pyaternev \\ 
   HSE University \\
   \And
   Anna Scherbakova \\
   HSE University \\
   \And
   Andrey Ustuzhanin \\
   HSE University \\
}

\begin{document}

\maketitle
\begin{abstract}
% The abstract paragraph should be indented 1/2~inch (3~picas) on both left and
% right-hand margins. Use 10~point type, with a vertical spacing of 11~points.
% The word \textsc{Abstract} must be centered, in small caps, and in point size 12. Two
% line spaces precede the abstract. The abstract must be limited to one
% paragraph.
A range of applications for automatic machine learning need the generation process to be controllable. In this work, we propose a way to control the output via a sequence of simple actions, that are called semantic code classes. Finally, we present  a semantic code classification task and discuss methods for solving this problem on the Natural Language to Machine Learning (NL2ML) dataset.
\end{abstract}
\section{Introduction}
\par Code generation has the potential to be useful in a range of applications, although there is a variety of limitations (\cite{DBLP:journals/corr/abs-2107-03374}). Despite the recent success in this area with Codex models (\cite{DBLP:journals/corr/abs-2107-03374}), controlled code generation remains challenging (\cite{DBLP:journals/corr/abs-2101-11149,DBLP:journals/corr/HuYLSX17}). Code generation models usually receive a code prompt (\cite{DBLP:journals/corr/abs-2002-08155, DBLP:journals/corr/abs-2107-03374} or a description of a task in natural language (\cite{DBLP:journals/corr/abs-2010-03150}), but the output is not controlled. 

\par Automatic machine learning (AutoML) is a specialized area of code generation. Controlled generation of machine learning (ML) code is crucial for AutoML applications because code requirements for the same task may vary drastically in practice. Some applications focus more on interpretability; others might be satisfied with uninterpretable models that perform well.

\par Many machine learning programs follow a specific sequence of steps. Such sequence is called an ML pipeline. Every machine learning program may be roughly split into code snippets, such that each snippet is responsible for a single small action. These small actions are called semantic code classes.

\par A sequence of semantic classes can control the generation process and provide a generative model with additional information about the requested pipeline. Such a sequence serves as a guide and indicates the most important steps.

\par Training a model to recognize a sequence of semantic actions requires a labeled dataset. The labeling task is highly specific and requires a lot of manual labor, hence it might be beneficial to train an additional model for labeling.

\par In this work, we present the semantic code classification task and compare the results of various models. Such representation allows to better understand the structure of the ML code and the contents of each step of possible pipelines.

% \par In this work, we propose an intermediate representation that might help with generating complete machine learning pipelines The proposed representation splits code into snippets so that each snippet is responsible for a single small action in the pipeline. These small actions are called semantic code classes. Such representation allows to better understand the structure of the ML code and the contents of each step of possible pipelines, which can contribute to the generation task. Also, we present a baseline for the semantic code classification task.

\section{Dataset}
\par Natural Language to Machine Learning (NL2ML) corpus is a collection of code blocks from Kaggle competition notebooks. It includes 101\,071 different notebooks comprising 2\,587\,074 cells (or code block), which participants submitted to 266 Kaggle competitions. 4\,748 of those code blocks have passed through manual assessment of its semantics. Each block is attributed to an upper-level category, which corresponds to a coarse meaning, and a second-level category, representing the finer semantic of that block. Publication of the corpus, along with the detailed description of the annotation process, is delayed due to legal considerations with Kaggle. 
\par The upper level categories attribute the general kind of code block. One block can belong to a single category. Here is the list of possible values:
\texttt{Hypothesis},
\texttt{Environment},
\texttt{Data extraction},
\texttt{Exploratory data analysis},
\texttt{Data transform},
\texttt{Model train},
\texttt{Model evaluation},
\texttt{Model interpretation},
\texttt{Hyperparameter tuning},
\texttt{Visualization}.
The lower-level category represents a more specific type, for example, upper-level class \texttt{Data transform} contains such items as \texttt{create dataframe}, \texttt{remove duplicates}, \texttt{correct missing values}, \texttt{feature engineering}, \texttt{filter} and others.
\par
\begin{figure}[h!]
	\centering
	\includegraphics[width=0.9\linewidth]{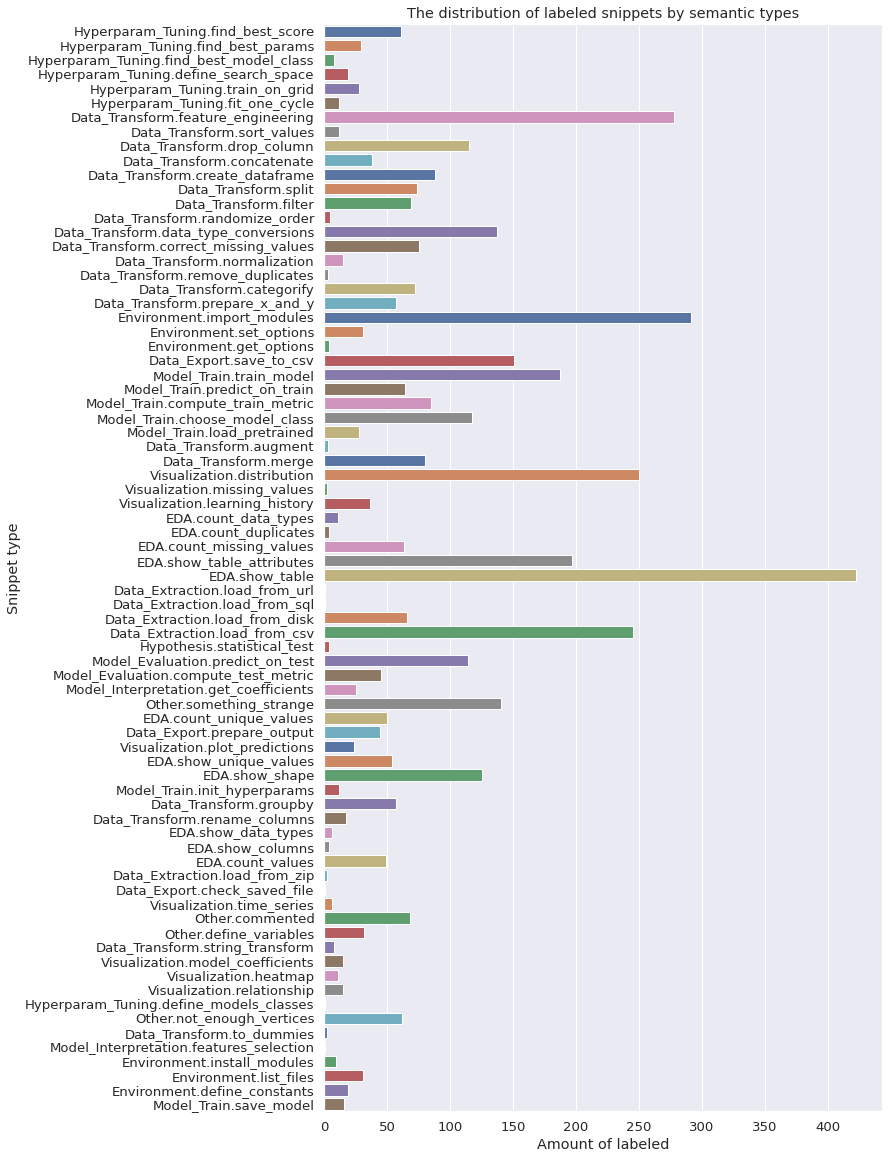}
    \caption{The distribution of labeled snippets by semantic types.}\label{fig:distr}
\end{figure}

 The dataset is very unbalanced (appendix \ref{fig:distr}). Some simple classes (\texttt{show table}) are overrepresented because data contains many similar examples. At the same time, many classes (\texttt{load data from zip}, \texttt{remove duplicates}) are represented by less than 10 examples. Another difficulty is that data is heavily skewed because a large portion of the snippets came from the same competition; hence the snippets in the same class tend to be similar in general.
\section{Code classification}
This section describes methods we used to classify code blocks into the lower level semantic classes. The details about metric computation are provided in the section \ref{sec:results}. For all models except neural networks we provide metrics with error bars that measure randomness in training the model.

The source code is available online\footnote{\url{https://github.com/whatevernevermindbro/nl2ml-mirror}}.

\subsection{Baseline solution: SVM}
Support vector machines classifier (\cite{bosertraining}) with linear kernel was used as baseline solution because the method does not require a large amount of samples. The multiclass support is handled according to a one-vs-one scheme (\cite{chang2011libsvm}). Baseline F1-score is 0.676 $\pm$ 0.029.

The F1-score is defined by the following formula:

$$F_1 = 2 \frac{precision \cdot recall}{precision + recall}$$

% All parameters were selected automatically.
% Standard deviation was calculated from cross-validation on 10 folds.

Due to the imbalance in the data, the F1-score was chosen as the target metric.

\subsection{Preprocessing}
All the code blocks were modified according to PEP8 using a standard Python utility, autopep8, to eliminate the problems with code-style differences.

Comments have been removed from the code. Despite the fact that they can contain a lot of information about the block's semantic type, their presence, as a rule, slightly degrades the quality of predictions. This is most likely because the content of the comments can be both useful and confusing. The text may contain information unrelated to the code block or unnecessary code that does not affect the semantic type.

For tokenization, the Byte-pair encoding method was used (\cite{DBLP:journals/corr/SennrichHB15}). The tokenizer was trained using unlabeled data. The size of the dictionary used is 30,000 tokens; during the training, the BPE-Dropout regularization technique was used (\cite{DBLP:journals/corr/abs-1910-13267}).

This tokenization has proven to be quite helpful for many models. The names of entities in the program code can consist of several words, written together or through an underscore \texttt{"\_"}. Splitting by the spaces cannot distinguish different words included in the names of entities, and breaking into n-grams may not be flexible since words in names can have different lengths. The BPE method was optimal because it can extract information about the donkeys that are often found in the names of entities.

Term frequency-inverse document frequency (TF-IDF) (\cite{ZHANG20112758}) was used as features for models. These statistics show the importance of the token in the document, given the context of the corpus.

\subsection{Naive Bayesian Classifier and  NB-SVM}
The Naive Bayesian Classifier (NB) (\cite{DBLP:journals/corr/Raschka14}) is a probabilistic model based on Bayes' theorem.
% Let $ y $ be the target variable, and $ x_1, \, \ldots, \, x_d $ be features. Then the probabilistic model approximates the conditional density:
% \[
% p\left(y\,|x_1,\,\ldots,\,x_d\right) = \frac{p\left(x_1,\,\ldots,\,x_d\,y\right)}{p\left(x_1,\,\ldots,\,x_d\right)}
% \]
NB Classifier is a common pre-neural network method for classifying text data. As a rule, it is assumed that the data has either a multinomial distribution or a Bernoulli distribution.

Since the dataset is unbalanced, estimates for the parameters of the multinomial distribution may be unstable, which negatively affects the quality of predictions. To combat this problem, a complementary NB was proposed in \cite{10.5555/3041838.3041916}.

The model with the Bernoulli distribution showed the best result among naive Bayesian classifiers. This is because this model works best with short texts (\cite{article}). Short texts predominate in the training dataset.

Metrics for NB classifier are in rows \texttt{Multinomial NB}, \texttt{Complementary NB}, \texttt{Bernoulli NB} of table \ref{table:all_results}.

NB classifier and SVM are often used as basic solutions in text classification problems. NB tends to perform better on shorter passages, while SVM works best on longer texts. The article \cite{wang-manning-2012-baselines} discusses the composition of an NB classifier and SVM, Naive Bayes - Support Vector Machine (NBSVM), which combines the advantages of both methods. The paper also states that binarization of NB features or using bigrams should improve results, but in our case it's the opposite. Metrics for NBSVM classifier are in rows \texttt{NBSVM}, \texttt{NBSVM (binarization)}, \texttt{NBSVM birams} of table \ref{table:all_results}.

\subsection{Neural networks}
Neural networks are an important class of machine learning algorithms. It is due to neural networks that the quality of text data processing has been improved. Unfortunately, many advanced neural network architectures require a lot of training data. There was not too much data in this case, but it was possible to obtain relatively good results.

There is not enough data to train vector representations, so frozen pretrained CodeBERT (\cite{DBLP:journals/corr/abs-2002-08155}) was used. CodeBERT has several drawbacks that are worth noting. Firstly, the important thing is that this model uses its own tokenizer. The tokens used do not quite match the machine learning code as they were derived from more general code, which means that machine learning-specific tokens may be too rare to enter the dictionary. Secondly, CodeBERT was trained on the program code not only in Python3 but also in several other languages: JavaScript, Java, PHP, Ruby, Go. Due to this, the connection of tokens with machine learning is becoming even less.

Another important point when training a neural network is the loss function, on which the success of the model depends a lot. The standard loss function in a classification problem is cross-entropy. Conventional cross-entropy is unstable to unbalanced data; the situation can be improved by adding weights for examples from different classes. Since there are a lot of classes, a vast number of hyperparameters appear, which is a big drawback since neural networks take a long time to learn.

For unbalanced data, there are alternative loss functions:
\begin{enumerate}
    \item Soft-F1. This loss function directly optimizes the target metric in the task at hand, which is a significant advantage. 
    
    The critical drawback of this loss function is that it requires a sizeable mini-batch size. If the batch size is too small, rare classes will often not be included in the batch, hence the precision and the recall are equal to 1, which is the optimal value for these metrics. If there are many such rare classes, the loss function will be extremely close to the optimum, leading to the fact that learning does not occur. In the dataset, quite a few classes contain less than 10 objects, which means that large batch size is required.
    \item Focal Loss. This loss function comes from computer vision (\cite{DBLP:journals/corr/abs-1708-02002}). The main idea is to adaptively reduce the weight of objects for which the model determines the class correctly and confidently. Due to this, it is possible to learn patterns even in rare classes.
\end{enumerate}

The following architectures were used:
\begin{enumerate}
    \item Recurrent neural network (RNN). GRU and LSTM were tried as a recurrent layer; the result is approximately the same. The recurrent network is bidirectional, since this configuration usually gives the best result, allowing more context for the tokens. Row \texttt{RNN} in results table (\ref{table:all_results}).
    \par The token sequence is fed into the inputs of the recurrent layer. The output goes through a dense layer, GELU activation, dropout, and outputs probabilities through dense layer and Softmax.
    \par Layer sizes are optimized with random search. The model with the best score uses a single LSTM layer with hidden state size of 180 and a hidden fully connected layer of size 269. The scheme is in appendix \ref{sec:rnn_arch}.
    \item RNN with additive attention (\cite{bahdanau2014neural}). The model differs from the usual RNN in only one layer: the attention layer added after the recurrent layer.
    \par The sizes of all layers were optimized using random search. The best model uses a single LSTM layer with hidden state size of 100. After attention, there is a hidden fully connected layer of size 135. The scheme of the model can be found in appendix \ref{sec:rnn_att_arch}.
    \par A layer of attention helps to identify the most important values in a sequence, which leads to a dramatic improvement in the result. Row \texttt{RNN+Attention} in results table (\ref{table:all_results}).
\end{enumerate}

\subsection{Augmentation}
Since the marked-up data is insufficient, augmentation was used. However, it is necessary that the code remains compilable and belongs to the same semantic type after modifications.

Many code blocks contain similar variable names (eg. \texttt{"df", "train", "test"}). They do not always provide additional information about the block type, but due to the frequency they are used by the classifier. Therefore, augmentation was applied, which masks some fraction of the variables in the block. the augmentation does not change the semantic of the block, as the names of user-defined variables can be almost anything. The variables in code were found using AST trees and replaced by masks, each variable in a block with a different one.

The augmentation was used with SVM and  Attention RNN. Results are in rows \texttt{SVM+Augmentation}, \texttt{RNN+Attention+Augmentation} in table (\ref{table:all_results}). The results have not improved enough(the value is within error bars), so we do not use augmentations in other experiments.

\subsection{Hierarchy}
The structure by which the code snippets are classified is a two-level graph: the first is the general class of action, the second is what exactly is done in the block. Therefore, it was decided to implement this structure as a two-level classifier, where the upper vertex is predicted first, and then the lower by a separate classifier, already trained on specific objects belonging to the child vertices.

The advantage of considering the upper classes separately is that in contrast to the original dataset, each class is represented by at least several dozen objects. At the same time, most of the very small original classes go into such a classification together with other small classes. Thus, the determination of their upper class is a more realistic task. If we correctly guessed the upper class, then small objects are likely to have no large competitors. However, if the class was identified incorrectly in the first step, the subsequent classification is guaranteed to be inaccurate.

SVM was chosen as the lower classifier since the model demonstrated the best results in previous experiments, while the upper classifiers were
\begin{enumerate}
\item RNN containing a bidirectional LSTM, the vector representations  were taken from the frozen pre-trained CodeBERT (\cite{DBLP:journals/corr/abs-2002-08155}).
\item SVM 
\end{enumerate}

In table \ref{table:hier_results_lvl1} the metrics of the upper level classification are presented. SVM achieved sufficient results. The overall results of hierarchy classifier are in  \ref{table:all_results} - rows \texttt{RNN + SVM Hierarchy}, \texttt{SVM + SVM Hierarchy} - this method allowed to significantly improve metrics.

\begin{table}[ht]
\caption{Upper level prediction results.}
\centering
\begin{tabular}{l l l}
\toprule
& \multicolumn{2}{c}{Metrics} \\
\cmidrule(r){2-3}
Model & F1-score &   Accuracy \\ [0.5ex] 
\midrule
RNN & 0.759 & 0.773 \\ [1ex] 
\midrule
SVM & \textbf{0.937} & \textbf{0.938} \\ [1ex] 
\bottomrule
\end{tabular}
\label{table:hier_results_lvl1}
\end{table}

\subsection{Semi-supervised, pseudo labels}
Since the amount of labeled data is extremely small, it was decided to experiment with semi-supervised models that allow to use unlabeled data (\cite{DBLP:journals/corr/abs-1911-04252}).

First, we trained a classic SVM model on the marked-up dataset. It's predictions on unlabeled data were used as pseudo labels. Then another SVM model with separately selected hyperparameters was trained on pseudo-labeled data and a part of marked-up data. It was tested on the rest of the labeled data.

We tested this approach using 20\%, 40\%, and 100\% of unlabeled data.
The results are in rows  \texttt{Pseudo labels 20\%},  \texttt{Pseudo labels 40\%},  \texttt{Pseudo labels 100\%}  in table (\ref{table:all_results}).

\section{Results}
\label{sec:results}
We evaluated models on marked-up code snippets of the NL2ML dataset. Several approaches demonstrated some improvements in metrics from the baseline solution. All of the resulting models are compared in the table (\ref{table:all_results}) by F1-score and accuracy. The best results are achieved by hierarchy and pseudo-labels models. The first one utilizes the knowledge of the classes structure, and the second one uses a great amount of unlabeled data.

The target metrics were computed on a test dataset that contains 20 \% of all labeled data. For each model, we optimized hyperparameters with random search. For neural networks, we used a validation set that contains 20 \% of labeled data.  For other methods, we used cross-validation with 10 folds.

\begin{table}[ht]
\caption{Summary table of all the models and their results}
\centering
\begin{tabular}{l l l}
\toprule
& \multicolumn{2}{c}{Metrics} \\
\cmidrule(r){2-3}
Model  & F1-score &  Accuracy \\ [0.5ex] 
\midrule
SVM+Linear (Baseline) & 0.676 $\pm$ 0.029 & 0.678 $\pm$ 0.031 \\
\hline
SVM + Poly & 0.620 $\pm$ 0.033 & 0.627 $\pm$ 0.030 \\
SVM + RBF & 0.672 $\pm$ 0.031 & 0.676 $\pm$ 0.031 \\
\midrule
Multinomial NB & 0.561 $\pm$ 0.031 & 0.583 $\pm$ 0.031\\
Complementary NB & 0.576 $\pm$ 0.024 & 0.587 $\pm$ 0.023  \\
Bernoulli NB & 0.609 $\pm$ 0.021 & 0.623 $\pm$ 0.026 \\
\midrule
NBSVM & 0.694 $\pm$ 0.013 & 0.704 $\pm$ 0.015 \\
NBSVM (binarization) & 0.684 $\pm$ 0.016 & 0.694 $\pm$ 0.015 \\
NBSVM (bigrams) & 0.689 $\pm$ 0.14 & 0.697 $\pm$ 0.14 \\
\midrule
RNN & 0.541 & 0.556\\
RNN + Attention & 0.639 & 0.648\\
\midrule
SVM + Augmentation & 0.686 $\pm$ 0.031 & 0.696 $\pm$ 0.029 \\
RNN + Attention + Augmentation & 0.638 & 0.652 \\
\midrule
RNN + SVM Hierarchy & 0.291 & 0.282 \\
SVM+SVM Hierarchy & 0.741 & 0.744 \\
\midrule
Pseudo labels 20 \% & 0.713 $\pm$ 0.014 & 0.724 $\pm$ 0.014 \\
Pseudo labels 40 \% & 0.727 $\pm$ 0.016 & 0.738 $\pm$ 0.014 \\
Pseudo labels 100 \% & \textbf{0.755} $\pm$ 0.03 & \textbf{0.764} $\pm$ 0.028\\
\bottomrule
\end{tabular}
\label{table:all_results}
\end{table}

% for presentation

% \begin{table}[ht]
% \caption{Summary table of all the models and their results}
% \centering
% \begin{tabular}{l l l}
% \toprule
% & \multicolumn{2}{c}{Metrics} \\
% \cmidrule(r){2-3}
% Proportion of pseudo-labeled data  & F1-score &  Accuracy \\ [0.5ex] 
% \midrule
% 20 \% & 0.713 $\pm$ 0.014 & 0.724 $\pm$ 0.014 \\
% 40 \% & 0.727 $\pm$ 0.016 & 0.738 $\pm$ 0.014 \\
% 100 \% & \textbf{0.755} $\pm$ 0.03 & \textbf{0.764} $\pm$ 0.028\\
% \bottomrule
% \end{tabular}
% \label{table:all_results}
% \end{table}

\section{Conclusion}
This paper compares methods for semantic classification of code snippets from NL2ML corpus, a large-scale dataset of data science-specific code harvested from Kaggle, the largest platform for data science competitions. The dataset includes only 4\,748 manually labeled snippets. 

We provided detailed qualitative and quantitative comparisons of the used classification methods. The final proposed model demonstrated better results than the baseline SVM solution, which has an F1-score of 0.67. The method with the best F1-score of 0.755 is an SVM trained on pseudo-labels. % The demonstrated results are significantly better than those obtained by solutions without Machine Learning, which only achieve an F1-score of $ \approx 0.4 $.

The proposed method allows to automatically markup a program code corpus with better accuracy and performance, as well as effectively assess the quality of data science code in Python. We also genuinely hope that this work can contribute to the development of the next generation of robust AutoML systems, as semantic classification provides an interlingua that can be used for translating between natural and programming languages.

% мы рассмотрели такую-то важную задачу на примере таких-то данных, и наши методы работают на этих данных лучше, чем бейзлайны на столько-то. ожидаем, что от применения наших методов в таких-то задачах станет намного лучше

% наша задача это обработка и понимание программного кода

%  В дальнейшем поможет сократить затраты на разметку
%  можно оценить произвольный код студента на питоне

\bibliography{neurips}
\bibliographystyle{neurips}

\appendix

\newpage

\section{Distribution of snippets by semantic types}
\label{sec:snippet_distr}

\newpage
\section{Architectures of neural nets}

\begin{figure}[h!]
	\centering
	\includegraphics[width=0.65\linewidth]{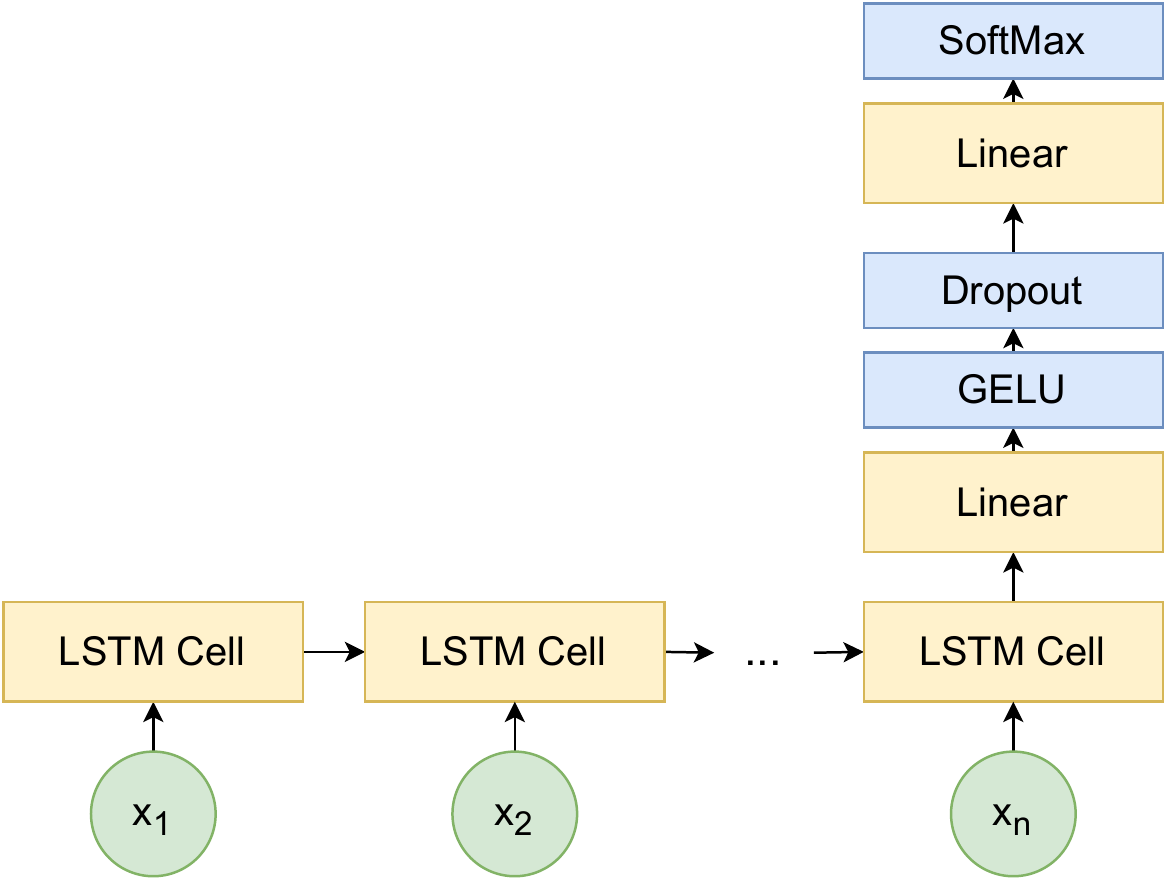}
	\caption{Architecture of recurrent neural network}
	\label{sec:rnn_arch}
\end{figure}

\begin{figure}[h!]
	\centering
	\includegraphics[width=0.65\linewidth]{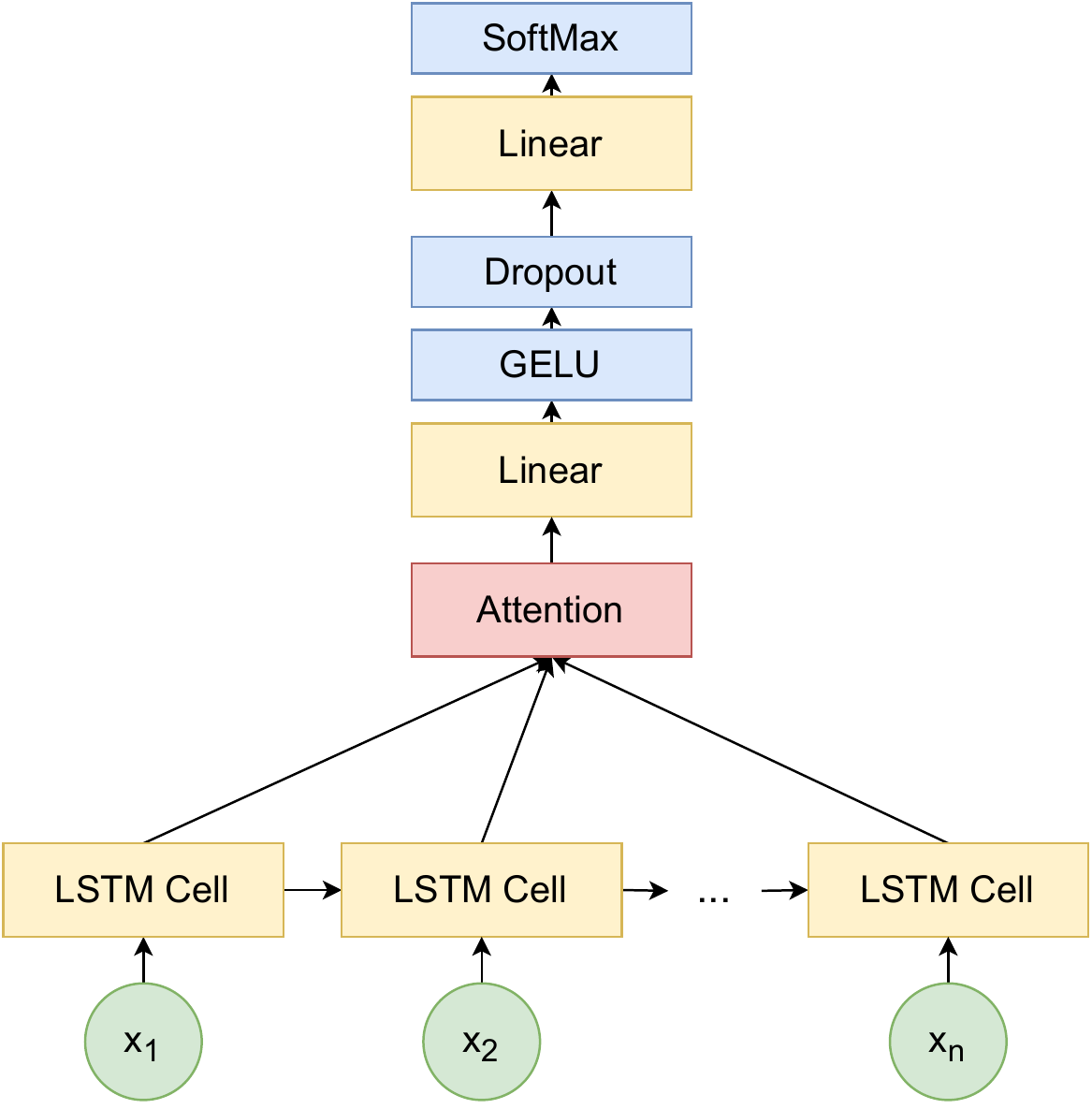}
	\caption{Architecture of recurrent neural network with attention}
	\label{sec:rnn_att_arch}
\end{figure}

\newpage
\section{Code snippets}
\label{sec:code_snippets}

\begin{longtable}{l l}
\caption{Examples of code snippets from NL2ML dataset} \\

\toprule
\textbf{Semantic class} & \textbf{Example} \\
\midrule
Data\_Transform.drop\_column          &  \begin{lstlisting}[language=Python]
train_df.drop("Date", inplace=True, axis=1)
test_df.drop("Date", inplace=True, axis=1)
\end{lstlisting} \\
\midrule
Model\_Train.choose\_model\_class & \begin{lstlisting}[language=Python]
from sklearn import linear_model

reg_CC = linear_model.Lasso(alpha=0.1)
reg_Fat = linear_model.Lasso(alpha=0.1)
\end{lstlisting} \\
\midrule
Hyperparams.define\_search\_space & \begin{lstlisting}[language=Python]
parameters = {'lstm_nodes': [14,16,20],
              'nb_epoch': [50],
              'batch_size': [32],
              'optimizer': ['adam']}
\end{lstlisting} \\
\midrule
Visualization.distribution & \begin{lstlisting}[language=Python]
fig = plt.figure()
fig.suptitle("Algorithm Comparison")
ax = fig.add_subplot(111)
plt.boxplot(results)
ax.set_xticklabels(names)
plt.show()
\end{lstlisting} \\
\midrule
Data\_Transform.normalization & \begin{lstlisting}[language=Python]
scaler = MinMaxScaler()
df['revenue'] = scaler.fit_transform(
    df[['revenue']]
)
\end{lstlisting} \\
\midrule
Model\_Train.train\_model & \begin{lstlisting}[language=Python]
rfr = RandomForestRegressor(
    n_estimators=200, 
    max_depth=5, 
    max_features=0.5, 
    random_state=449,
    n_jobs=-1
)
rfr.fit(x_train, y_train)
\end{lstlisting} \\
\bottomrule
\end{longtable}

\newpage
\section{Selected hyperparameters for each model}

\begin{longtable}{ l l r r } % |p{105mm}|
\caption{Hyperparameters for models without hierarchy and first level of hierarchical models} \\

\toprule
\textbf{Model} & \textbf{Hyperparameter} & \textbf{Type} & \textbf{Value} \\
\midrule
SVM+Linear (Baseline) & C & numeric & 37.17 \\
                      & min\_df for TF-IDF & integer & 2  \\
                      & max\_df for TF-IDF & numeric & 0.31 \\
\midrule
SVM + Poly & C & numeric & 1.43 \\
           & Degree of poly kernel & integer & 3 \\
           & min\_df for TF-IDF & integer & 6  \\
           & max\_df for TF-IDF & numeric & 0.30 \\
\midrule
SVM + RBF & C & numeric &  8.71 \\
              & min\_df for TF-IDF & integer &  7  \\
              & max\_df for TF-IDF & numeric & 0.39 \\
\midrule
Multinomial NB & Alpha & numeric & 0.01 \\
               & min\_df for TF-IDF & integer & 2 \\
               & max\_df for TF-IDF & numeric & 0.57 \\
\midrule
Complementary NB & Alpha & numeric & 0.29 \\
                 & min\_df for TF-IDF & integer & 2 \\
                 & max\_df for TF-IDF & numeric & 0.30 \\
\midrule
Bernoulli NB & Alpha & numeric & 0.01 \\
             & min\_df for TF-IDF & integer & 5 \\
             & max\_df for TF-IDF & numeric & 0.30 \\
\midrule
NBSVM  & C for SVM & numeric & 0.50 \\
       & Kernel type & categorical & Linear \\
       & Alpha for Naive Bayes & numerical & 4.22 \\
       & min\_df for TF-IDF & integer & 1 \\
       & max\_df for TF-IDF & numeric & 0.94 \\
\midrule
NBSVM (binarization) & C for SVM & numeric & 0.10 \\
                     & Kernel type & categorical & Linear \\
                     & Alpha for Naive Bayes & numerical & 0.28 \\
                     & min\_df for TF-IDF & integer & 3 \\
                     & max\_df for TF-IDF & numeric & 0.94 \\
\midrule
NBSVM (bigrams) & C for SVM & numeric & 0.10 \\
                & Kernel type & categorical & Linear \\
                & Alpha for Naive Bayes & numerical & 2.51 \\
                & min\_df for TF-IDF & integer & 2 \\
                & max\_df for TF-IDF & numeric & 0.95 \\
\midrule
RNN & RNN cell type & categorical & LSTM \\
    & Size of hidden state & integer & 180 \\
    & Size of linear layer & integer & 269 \\
\midrule
RNN + Attention & RNN cell type & categorical & LSTM \\
                & Size of hidden state & integer & 100 \\
                & Size of linear layer & integer & 135 \\
\midrule
SVM + Augmentation & C & numeric & 5.82\\
                   & Kernel type & categorical & Linear\\
                   & min\_df for TF-IDF & integer & 2\\
                   & max\_df for TF-IDF & numeric & 0.30 \\
                   & Percent of masked variables  & numeric & 0.93 \\
\midrule
RNN + Attention + Augmentation & RNN cell type & categorical & LSTM \\
    & Size of hidden state & integer & 169 \\
    & Size of linear layer & integer & 255 \\
    & Percent of masked variables  & numeric & 99.7\% \\
\midrule
SVM+SVM Hierarchy (1st level) & C & numeric & 149.65 \\
                              & Kernel type & categorical & Poly \\
                              & Degree of poly kernel & integer & 2 \\
\midrule
RNN+SVM Hierarchy (1st level) & RNN cell type & categorical & LSTM \\
                              & Size of hidden state & integer & 143 \\
                              & Size of linear layer & integer & 185\\
\midrule

Pseudo labels 20 \% & C & numeric & 98.37 \\
                    & Kernel type & categorical & linear\\
                    & min\_df for TF-IDF & integer & 3  \\
                    & max\_df for TF-IDF & numeric & 0.53 \\
\midrule
Pseudo labels 40 \% & C & numeric & 121.59 \\
                    & Kernel type & categorical & linear\\
                    & min\_df for TF-IDF & integer & 3  \\
                    & max\_df for TF-IDF & numeric & 0.41 \\
\midrule
Pseudo labels 100\% & C & numeric & 145.56 \\
                    & Kernel type & categorical & linear\\
                    & min\_df for TF-IDF & integer & 2  \\
                    & max\_df for TF-IDF & numeric & 0.26 \\
\bottomrule
\end{longtable}

\begin{longtable}{ l l r r } % |p{105mm}|
\caption{Hyperparameters for SVM+SVM Hierarchy (2nd level). Separate parameters for each upper level class. If upper level class is not present in this table, then it has one subclass and hence does not require a second level classificator.} \\

\toprule
\textbf{Upper level class} & \textbf{Hyperparameter} & \textbf{Type} & \textbf{Value} \\
\midrule
Data Extraction & C & numeric & 135.49 \\
                & Kernel type & categorical & Linear \\ 
                & min\_df for TF-IDF & integer & 50 \\ 
                & max\_df for TF-IDF & numeric & 0.89 \\
\midrule
EDA & C & numeric & 4.70 \\
    & Kernel type & categorical & Linear \\
    & min\_df for TF-IDF & integer & 4 \\
    & max\_df for TF-IDF & numeric & 0.61 \\
\midrule
Environment & C & numeric & 232.06 \\
            & Kernel type & categorical & Linear \\
            & min\_df for TF-IDF & integer & 6 \\
            & max\_df for TF-IDF & numeric & 0.81 \\
\midrule
Model\_Train & C & numeric & 9.63 \\
             & Kernel type & categorical & RBF \\
             & min\_df for TF-IDF & integer & 17 \\
             & max\_df for TF-IDF & numeric & 0.47 \\
\midrule
Data\_Transform & C & numeric & 298.83 \\
                & Kernel type & categorical & Linear \\
                & min\_df for TF-IDF & integer & 2 \\
                & max\_df for TF-IDF & numeric & 0.20 \\
\midrule
Model\_Evaluation & C & numeric & 0.84 \\
                  & Kernel type & categorical & Poly \\
                  & Degree of poly kernel & integer & 3 \\
                  & min\_df for TF-IDF & integer & 16 \\
                  & max\_df for TF-IDF & numeric & 0.50 \\
\midrule
Data\_Export & C & numeric & 5.35 \\
             & Kernel type & categorical & RBF \\
             & min\_df for TF-IDF & integer & 23 \\
             & max\_df for TF-IDF & numeric & 0.83 \\
\midrule
Hyperparam\_Tuning & C & numeric & 20.03 \\
                   & Kernel type & categorical & RBF \\
                   & min\_df for TF-IDF & integer & 6 \\
                   & max\_df for TF-IDF & numeric & 0.26 \\
\midrule
Visualization & C & numeric & 865.58 \\
                & Kernel type & categorical & Linear \\
                & min\_df for TF-IDF & integer & 2 \\
                & max\_df for TF-IDF & numeric & 0.66 \\
\bottomrule
\end{longtable}

\begin{longtable}{ l l r r } % |p{105mm}|
\caption{Hyperparameters for RNN+SVM Hierarchy (2nd level). Separate parameters for each upper level class. If upper level class is not present in this table, then it has one subclass and hence does not require a second level classificator.} \\

\toprule
\textbf{Upper level class} & \textbf{Hyperparameter} & \textbf{Type} & \textbf{Value} \\
\midrule
Data Extraction & C & numeric & 7.66 \\
                & Kernel type & categorical & RBF \\ 
                & min\_df for TF-IDF & integer & 17 \\ 
                & max\_df for TF-IDF & numeric & 0.88 \\
\midrule
EDA & C & numeric & 1.71 \\
    & Kernel type & categorical & Linear \\
    & min\_df for TF-IDF & integer & 6 \\
    & max\_df for TF-IDF & numeric & 0.32 \\
\midrule
Environment & C & numeric & 462.65 \\
            & Kernel type & categorical & RBF \\
            & min\_df for TF-IDF & integer & 16 \\
            & max\_df for TF-IDF & numeric & 0.87 \\
\midrule
Model\_Train & C & numeric & 4.58 \\
             & Kernel type & categorical & Linear \\
             & min\_df for TF-IDF & integer & 1 \\
             & max\_df for TF-IDF & numeric & 0.90 \\
\midrule
Data\_Transform & C & numeric & 204.62 \\
                & Kernel type & categorical & Linear \\
                & min\_df for TF-IDF & integer & 2 \\
                & max\_df for TF-IDF & numeric & 0.45 \\
\midrule
Model\_Evaluation & C & numeric & 2.03 \\
                  & Kernel type & categorical & Linear \\
                  & min\_df for TF-IDF & integer & 3 \\
                  & max\_df for TF-IDF & numeric & 0.63 \\
\midrule
Data\_Export & C & numeric & 26.49 \\
             & Kernel type & categorical & Poly \\
             & Degree of poly kernel & numeric & 2 \\
             & min\_df for TF-IDF & integer & 23 \\
             & max\_df for TF-IDF & numeric & 0.93 \\
\midrule
Hyperparam\_Tuning & C & numeric & 14.58 \\
                   & Kernel type & categorical & Linear \\
                   & min\_df for TF-IDF & integer & 10 \\
                   & max\_df for TF-IDF & numeric & 0.65 \\
\midrule
Visualization & C & numeric & 2.48 \\
                & Kernel type & categorical & Linear \\
                & min\_df for TF-IDF & integer & 2 \\
                & max\_df for TF-IDF & numeric & 0.21 \\
\bottomrule
\end{longtable}

\end{document}